\documentclass{article}
\usepackage[utf8]{inputenc}
\usepackage{graphicx} 
\usepackage[english]{babel}

\title{Explainable artificial intelligence (XAI), the goodness criteria and the grasp-ability test}

\author{Tae Wan Kim \\ Carnegie Mellon University
\texttt{twkim@andrew.cmu.edu}}

\begin{document}
\maketitle

\begin{abstract}
This paper introduces the ``grasp-ability test'' as a ``goodness'' criteria by which to compare which explanation is more or less meaningful than others for users to understand the automated algorithmic data processing.
\end{abstract}

\section{Why XAI?}
\begin{figure}
\centering
  \includegraphics[width=.7\linewidth]{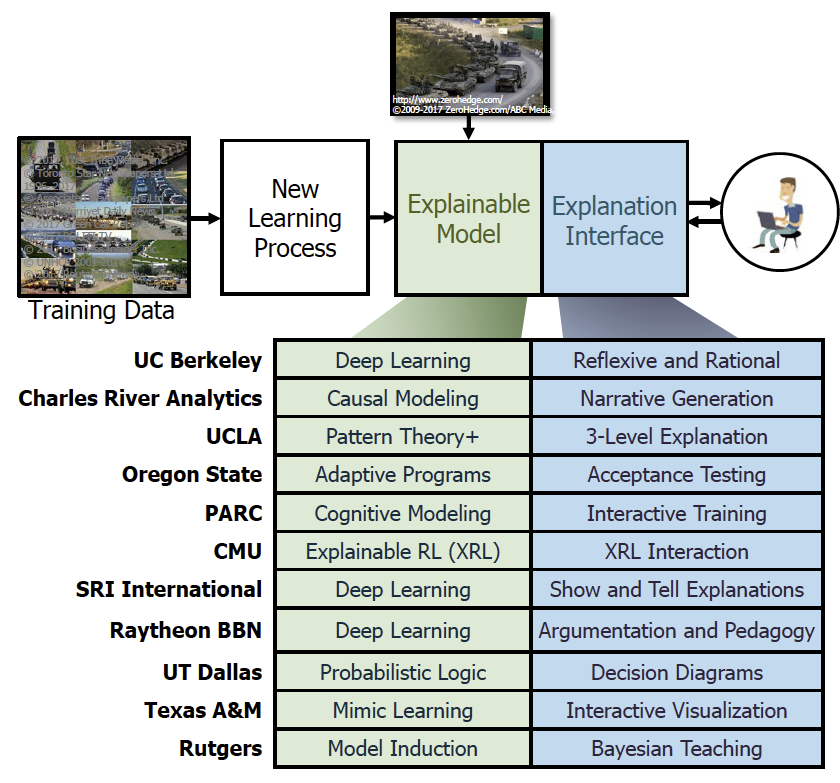}
  \caption{11 U.S. research labs funded by DARPA to develop XAI\protect\cite{gunning2017explainable}}
  \label{darpa}
\end{figure}

A growing number of researchers attempt to develop explainable AIs (hereafter, XAI) to meet practical (e.g., explainability is positively correlated to users' learning performance), legal (e.g., explainability is required for S.E.C. to scrutinize AI-powered trading techniques; liability issues) and ethical expectations (e.g., right to explanation; trust; autonomy). Different researchers use different ideas of what an explanation is \cite{lipton2016mythos}. For example, as Figure \ref{darpa} shows, 11 U.S. research groups, funded by DARPA, are currently developing XAI in different manners.

Then, a question is raised: how can we know which model of XAI is good enough or better/worse than others? To answer this, we need a ``goodness'' criteria. But the question of the goodness criteria has been under-explored. In this paper, we develop a conceptual framework of the ``goodness'' criteria. 

\section{Correct or good explanation?}
A useful place to begin with is the philosophy of science literature, wherein philosophers have studied scientific explanation since 1960s \cite{sep-scientific-explanation}. Largely, theories of explanation can be distinguished into two categories and the two kinds can be generically expressed as the following forms (E means ``explanation''; q means ``why-question.''):

\begin{itemize}
    \item Non-pragmatic theory of explanation: ``E is the \textit{correct} answer of q.''
    \item Pragmatic theory of explanation: ``E is a \textit{good} answer for an ex-plainer to give in answering q to an audience.''
\end{itemize}

The most significant difference between the two models is that there is no space for the audience in non-pragmatic theories. To clarify the difference, suppose that a serious bio-medical researcher is asked by a kindergartner to answer why his older brother has Down's Syndrome. The researcher offers the boy a latest scientific research paper about how the number of chromosomes 21 is related to Down's Syndrome, etc. The boy cannot understand the answer. The content in the paper is a correct explanation and it is the boy who does not get that. The correctness of the answer has nothing to do with whether the boy is capable of understanding it or not. Non-pragmatic theories are typically, but not always, synced with ``explanation monism''---a position that there is only one true explanation.

However, pragmatic theories of explanation submit that a correct explanation is not necessarily a good explanation. Consider how you would explain evolution to a kindergartner. A workable resource is to offer a visualization showing that humans descended from apes. But this explanation is not a correct answer, strictly speaking, because humans and apes descended from a common hominid ancestor which was not an ape. Thus, according to a non-pragmatic theory of explanation, the visualization fails to constitute an explanation. But it might be true that the perfectly correct explanation is not as useful to help a kindergartner understand evolution as the moderately, but not perfectly, accurate explanation. 

Pragmatic theorists submit that there is no point of offering an explanation if the explanation is not to be understood by the listener, so they argue that the definition of an explanation should necessarily have a place for the listener. Since different listeners have different knowledge bases, pragmatic theories are naturally allied with ``explanation pluralism''---a position that a phenomenon can have multiple explanations. Recently, pragmatic theories and explanation pluralism have started gaining much attraction in philosophy science. 

The first thesis of this paper is that when it comes to XAI, pragmatic theories of explanation are more appropriate than non-pragmatic theories. The practical, legal and ethical demands for companies and researchers to develop XAI largely come from the expectations that human users legitimately have. What is the point of XAI if human users do not understand that? 

A thorny problem that any pragmatic theories of explanation inherently have is that spectacularly inaccurate or deceptive explanations might be useful to users, depending upon the users' goals.  If the objective function is simply to make more oncologists to trust AI's diagnostic predictions, Why not offering a hypnotizing explanation to the surgeons? How can we overcome this problem? One solution is to go back to the non-pragmatic theory. But we cannot go back to non-pragmatic theories given the importance of human users' understanding in XAI. The question is, then, how to develop a pragmatic framework of explanation that is reasonably non-subjective. The second thesis of this paper is that it is very possible to develop a non-subjective pragmatic framework of explanation, which we will develop in the next section. 

\section{Introducing the grasp-ability test}
To reformulate the pragmatic model of explanation to be non-subjective, we draw upon the literature about relationships between ``understanding'' and explanation in philosophy of science \cite{grimm2016explaining}. In particular, we rely on and further extend ``the ability theory of understanding,'' according to which, roughly speaking, the speaker well explains to the extent that the hearer understands the explanation\cite{woodward2005making, grimm2016explaining, hills2013moral, hills2016understanding}. This perspective is appropriate for our context, because no matter how scientifically rigorous an explanation of how an ML makes a decision is, if users have difficulties understanding that, that is not a desirable form of explanation in the context of XAI.

Here’s our definition of a good explanation in AI/ML algorithms, in the case of the ``black-box'' model (e.g., deep learning):
\begin{itemize}

   \item The grasp-ability test: Explanation E is good to the extent that Audience S is able to understand that ``x because y'' (e.g., = there is a correlation between x and y and your x happens to be very high), where S understands that ``x because y'' to the extent that 
   
\begin{itemize}
  
    \item the counter-factual condition: S is able to answer context-relevant ``What-if-things-had-been-different?'' questions; 
    
    \item the moderate factative condition: propositions (or non-propositional representations such as diagrams, graphs, etc.) constituting ``x because y'' are reasonably accurate; and,
    
    \item the no-luck condition; S's ability is inconsistent with luck (A does not luckily answer counter-factual questions).

\end{itemize} 
\end{itemize}

Explanations can be propositional or non-propositional (e.g., visualization; using images of a planetarium to explain retrograde motion). If there are domains in which non-propositional representations turn out to be a better form of resources by which the audience is able to answer counter-factual questions, non-propositional resources are more appropriate in that domain.

\subsection{The counter-factual condition} The crucial element of our definition is to define explanation through understanding, and, in turn, understanding through audience's \textit{ability} to answer counter-factual questions. Why is the ability-giving component is important part of good explanation? Here is an illuminating example, although not directly about AI:

\begin{quote}
   Suppose that a climate scientist explains to her young son that the global mean surface temperature has massively increased since the middle of the 20th century because of increasing greenhouse gas concentrations. Since she is right and her son has good reasons to believe her explanation, he may be said to know why the global mean temperature has increased. But he does not seem to understand why. When asked why this is so, all he can do is to repeat his mother’s explanation. The problem seems to be that he does not really grasp the explanation. But what exactly is he lacking? (p. 12)\cite{grimm2016explaining}
\end{quote} 

The young son does not grasp the explanation because the explanation lacks a conception of how increasing concentrations of greenhouse gases leads to global warming. That is, one's grasping an explanation requires some conception about underlying mechanism about x because y. Once the young son grasps the underlying mechanism, he can grasp the whole explanation about the global warming. 

Now suppose that the young son's mother asks, ``Do you understand the global warming?'' He replies, ``Yes. Your explanation was good.'' But his answer may be too subjective. How can the mother objectively know that his son really understands the explanation so that her explanation was good? The example above already signifies that understanding is best captured by an ability to answer counter-factual questions. Once the son grasps the underlying mechanism (i.e., the greenhouse effect), he can answer counter-factual questions. For instance, ``What if concentrations of greenhouse gases did not increase?'' If he answered, ``The global mean temperature wouldn't increase dramatically'' it can be tentatively said that he grasps the explanation and the explanation is good \footnote{This hypothetical scenario is based on \cite{bostrom2014ethics}}.

\begin{figure}
    \centering
    \includegraphics[width=.7\linewidth]{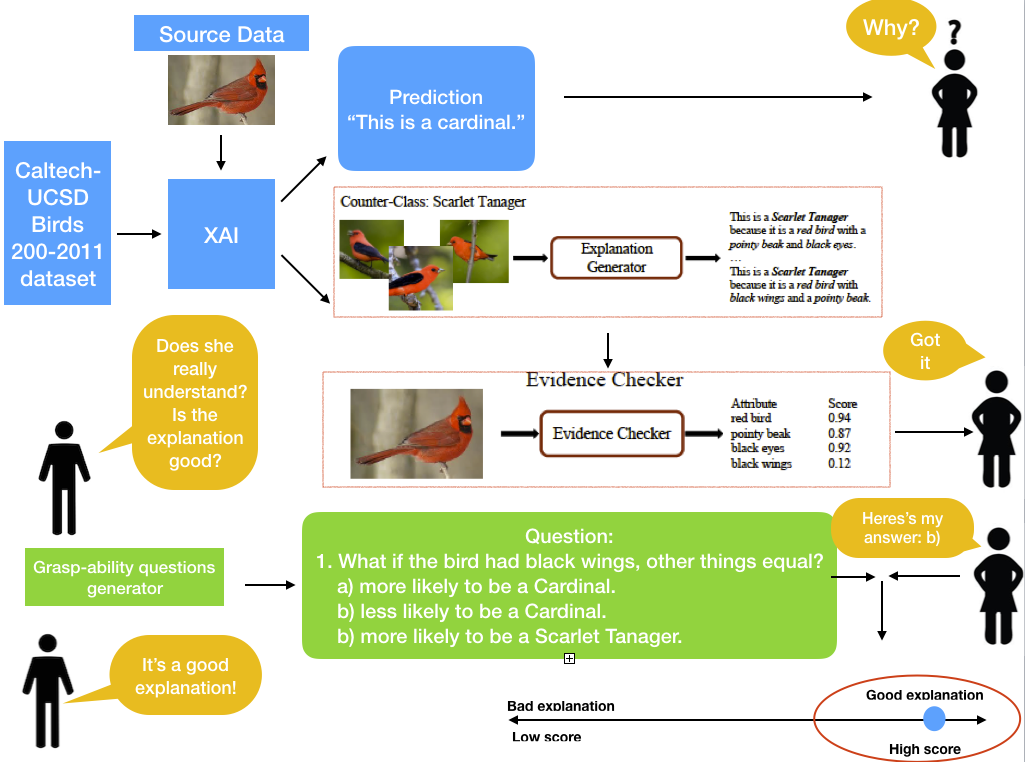}
    \caption{An example of the grasp-ability test 1}
    \label{bird test}
\end{figure}

Now move on to an hypothetical XAI case that I create based on an existing XAI work \cite{hendricks2018generating} (see Figure \ref{bird test}). Hendricks et al. use ``rejections of alternative choices'' to give an explanation. Their XAI attempts to explain why an image of a bird (e.g., cardinal) is classified as a ``cardinal'' by rejecting an alternative, a ``Scarlet Tanager,' which has similar features as cardinal except for black wings. The audience can understand that the algorithm learned that Scarlet Tanagers is a red bird with a pointy beak, black eyes and black wings, whereas cardinals have the same features except for black wings, and the image of the bird in input does not have black wings, so the algorithm classified it as a `scarlet.'  

The question is: How can we know whether the explanation is a good and meaningful one? A relevant counter-factual question in this context is ``What if the bird had black wings, other things equal?" Suppose that the counter-factual question is a multiple-choice question and it has three options: (a) more likely to be a cardinal; (b) less likely to be a cardinal; (c) more likely to be a Scarlet Tanager. The correct answer is (b). The explanation is good from a pragmatic point of view only if the audience can confidently chooses (b).

Now move on to another hypothetical XAI case: You are a person from a racially underrepresented group, say black, and you recently applied to an online mortgage approval system and were rejected.  The bank that hosts the online application system has recently started using AI to recommend mortgage applications for approval.  You happened to know that the bank’s approval rate for clients of the same race as yours has recently abnormally decreased, for no good reason.  You meet with a representative of the bank and claim that the bank has racially discriminated against you, and that the bank should be held liable for the discrimination (e.g., remedying the problem, compensating for any physical or psychological losses, if any, and taking action to prevent similar problems in the future). 

To find out why, the bank representative submits ten fake applications equally qualified as yours (as judged by independent human evaluators) that consist of 5 whites and 5 blacks. The AI accepts all white applicants but only 2 black. After some further investigation, the bank learns that the AI had learned that African American names are highly correlated with the zip codes where poverty rate is high and the machine realized the high correlation between poverty rate and the repayment rate. The bank now offers the above discovery as an explanation to the applicant. 

\begin{figure}
    \centering
    \includegraphics[width=.7\linewidth]{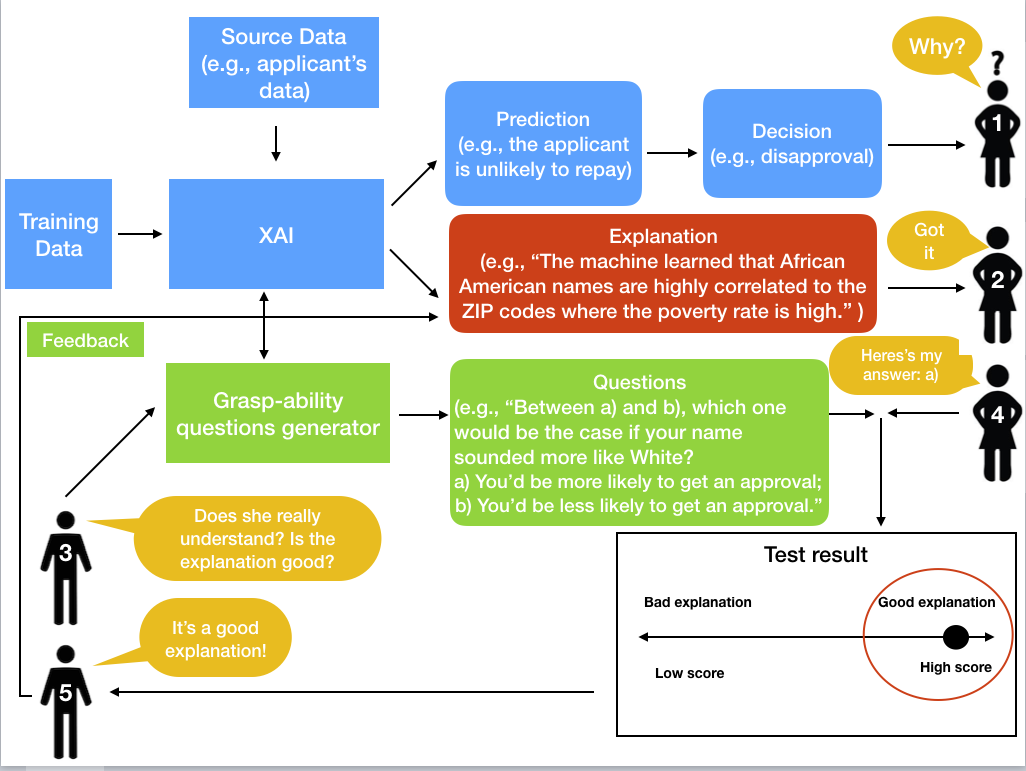}
    \caption{An example of the grasp-ability test 2}
    \label{race}
\end{figure}

The question is: How can we know whether the explanation is a good and meaningful one? A relevant counter-factual question in this context is ``What if your name sounded more like a White?'' (see figure \ref{race}) Suppose that the counter-factual question is a multiple-choice question and it has two options: (a) You'd be more likely to get an approval; (b) you'd be less likely to get an approval. An explanation is good only if the applicant confidently chooses (a). 

Of course, by going through the process, both the bank and the applicant can understand that the AI illegitimately discriminated against the applicant.

\subsection{The moderate factative condition} The propositional or non-propositional resources/representations must be reasonably accurate---namely, representations correspond to the internal workings of the AI.  But the factative condition does not require all resources are strictly true. Reconsider how you would explain evolution to a kindergartner. 

\subsection{The no-luck condition} It is possible for the young son or the applicant above to luckily answer the counter-factual question. To distinguish cases of epistemic luck from genuine grasping, the audience's ability to answer relevant counter-example questions must be further verified by, perhaps, variations of the same question or several different questions. If the audience answered all the questions correctly and consistently, the ability to answer is high, so the given explanation is good, and \textit{vice versa}.

\section{Concluding remarks}
Let us dub the whole process of finding out whether the audience understands an explanation through the three conditions above the ``grasp-ability'' test. A good explanation is not necessarily perfectly transparent, nor scientifically rigorous. A good explanation is a pragmatic resource by which to maximize the grasp-ability score of the audience. Thus, in theory, we can think about a grasp-ability test to compare which explanation is better than others. In addition, regulators can set up a domain-specific minimum score that any XAI should pass. There is no way to determine the minimum score requirement without any context. Perhaps, a similar notion is the idea that an alpha of 0.05 is used as the cutoff for significance in various sciences but the p-value is a human consensus rather than a natural value, but p-value test is still an useful test.

\bibliographystyle{unsrt}
\bibliography{ex}

\end{document}